\documentclass[final,3p,times]{elsarticle}
\usepackage{amssymb}
\usepackage{amsmath}
\usepackage[T1]{fontenc}
\usepackage{mathtools}         
\usepackage{amsfonts}
\usepackage{mathrsfs}
\usepackage[bold]{hhtensor}
\usepackage{algorithm}
\usepackage{algpseudocode}
\usepackage{enumitem}
\usepackage{booktabs}
\usepackage{threeparttable}
\usepackage{comment}
\usepackage{placeins}

\usepackage{url}
\usepackage{xcolor}

\usepackage{graphicx}
\usepackage{xcolor}
\usepackage{subcaption}

\usepackage{siunitx}
\usepackage{booktabs}
\usepackage{makecell}
\usepackage{multirow}
\usepackage{float}
\usepackage{placeins}
\usepackage{fancyhdr}  
\usepackage{tikz}
\usepackage[nolist]{acronym}

\usepackage{acronym}
\usepackage{pifont}
\usepackage{lineno}
\usepackage{geometry}
\usepackage[colorlinks=true, linkcolor=blue, citecolor=blue, urlcolor=blue]{hyperref}

\begin{acronym}
\acro{PDEs}[PDEs]{partial differential equations}
\acro{BC}{boundary condition}
\acro{KANs}{Kolmogorov-Arnold networks}
\acro{KAN}{Kolmogorov-Arnold network}
\acro{SciML}{Scientific machine learning}
\acro{BCs}{\ac{BC}s}
\acro{MLPs}{multilayer perceptrons}
\acro{ML}{machine learning}
\acro{DNN}{deep neural network}
\acro{RBFs}{radial basis functions}
\acro{PINNs}{physics-informed neural networks}
\acro{DeepONet}{deep operator networks}
\acro{EDNNs}{evolutionary deep neural networks}
\acro{EvoKAN}{evolutionary \ac{KAN}}
\acro{GP}{Gaussian process}
\acro{SAV}{scalar auxiliary variable}
\acro{EDNN}{evolutionary deep neural networks}
\acro{EvoKAN}{evolutionary \ac{KAN}}
\acro{PINN}{physics-informed neural networks}
\end{acronym}

\journal{Computer Methods in Applied Mechanics and Engineering}
\begin{document}
\begin{frontmatter}

\title{Low-Rank Evolutionary Deep Neural Networks via Adaptive Tangent-Space Reduction}

\author[label1]{Jiahao Zhang} %% Author name
\ead{zhan2296@purdue.edu}

\author[label2]{Shiheng Zhang}
\ead{shzhang3@uw.edu}

\author[label1,label3]{Guang Lin\corref{cor1}}
\cortext[cor1]{Corresponding author}
\ead{guanglin@purdue.edu}

%% Author affiliation
\affiliation[label1]{organization={Purdue University, School of Mechanical Engineering},%Department and Organization
            % addressline={585 Purdue Mall}, 
            city={West Lafayette},
            postcode={47906}, 
            state={IN},
            country={United States}}

\affiliation[label2]{organization={University of Washington, Department of Mathematics},
            % addressline={Mathematical Sciences Bldg, 150 N University St},
            city={Seattle},
            postcode={98195},
            state={WA},
            country={United States}}

\affiliation[label3]{organization={Purdue University, Department of Mathematics},
            % addressline={Mathematical Sciences Bldg, 150 N University St},
            city={West Lafayette},
            postcode={47906},
            state={IN},
            country={United States}}

%% Abstract
\begin{abstract}

Evolutionary deep neural networks (EDNNs) solve time-dependent partial
differential equations by evolving the neural-network parameters
sequentially in time through a local least-squares problem. Their main
computational bottleneck is that each time step requires the solution of a dense linear system whose dimension equals the total number of
trainable parameters. We propose a low-rank evolutionary deep neural
network (LR-EDNN) method that reduces this cost through adaptive
tangent-space projection. This construction replaces direct bilinear low-rank factor evolution by a linear reduced problem while preserving the sequential-in-time structure of EDNN. We construct the reduced Jacobian directly through layerwise Jacobian--vector products, without forming the full Jacobian. We further establish a finite-time comparison estimate: the deviation of the LR-EDNN trajectory from full EDNN is bounded by a discrete Gr\"onwall accumulation of the local tangent-space projection defects, with amplification governed by the assumed Lipschitz and directional-coercivity constants. Numerical experiments on a porous-medium
equation with drift, one- and two-dimensional Allen--Cahn equations, and two-dimensional viscous Burgers' equations demonstrate that LR-EDNN substantially reduces computational cost while maintaining the accuracy and qualitative fidelity of the full EDNN solver when the rank is chosen adequately.

\end{abstract}

%%Graphical abstract
%\begin{graphicalabstract}
%\includegraphics[width=1\linewidth]%{graphical_abstract.png}
%\end{graphicalabstract}

%%Research highlights
%\begin{highlights}
%\item We formulate low-rank EDNN as an adaptive projected sequential-in-time neural solver rather than as a bilinear low-rank factor update.
%\item The reduced update space is exactly the tangent space of a fixed-rank matrix manifold associated with the truncated layer weights.
%\item The reduced update is the best approximation of the full EDNN update in a Jacobian-induced seminorm, with a precise residual decomposition.
%\item Numerical results on nonlinear diffusion, phase-field, and convection-diffusion problems show substantial speedups while retaining accuracy at sufficiently large rank.
%\end{highlights}

%% Keywords

\end{frontmatter}

%% main text
\section{Introduction}

Deep learning has become an important computational paradigm for partial differential equations (PDEs), especially in high-dimensional, multiscale, or data-limited settings~\cite{han2018solving, ew2018deep}. Among neural PDE methods, two major viewpoints have been especially influential.

The first is the global-in-time residual-minimization viewpoint
represented by physics-informed neural networks (PINNs) and related
methods~\cite{raissi2019physics, luo2025physics, cai2021physics, mao2020physics, kissas2020machine, rao2021physics, haghighat2021physics, chen2020physics, lu2021physics, cai2021deepm, sun2022applications, wang2022and, wang2021understanding, ji2021stiff}. In this class of approaches, a single network is
trained over the full space-time domain, typically by minimizing a loss composed of PDE residuals, boundary conditions, and initial conditions. These methods are flexible and broadly applicable, but they often face optimization difficulties for long-time integration, stiff dynamics, and multiscale behavior.

The second is the operator-learning viewpoint, in which the goal is to
learn a map between function spaces rather than a single solution
trajectory~\cite{lu2021learning, li2020fourier, chen1995universal, sitzmann2020implicit, kashinath2020enforcing, li2020neural}. Operator-learning methods have shown impressive performance in many-query and parametric settings, but they generally depend on large collections of precomputed training pairs, whose generation may itself be computationally expensive.

For time-dependent PDEs, a third viewpoint is particularly appealing:
preserve the causal structure of the evolution problem and advance the
neural representation sequentially in time. Evolutionary deep neural
networks (EDNNs), introduced by Du and Zaki~\cite{du2021evolutional}, embody this idea by representing the spatial solution profile at each time level with a neural network whose parameters satisfy a local least-squares evolution problem. This sequential-in-time perspective aligns naturally with time-marching methods from numerical analysis and has already been shown to perform well for nonlinear and long-time dynamics~\cite{kim2025multi, zhang2024energy}.

From a broader perspective, EDNN belongs to the family of neural
Galerkin-type methods, in which the key computational task at each time step is to compute a parameter velocity whose induced solution
derivative best matches the governing PDE under the current network
representation~\cite{bruna2024neural, berman2024neural}. This viewpoint immediately exposes the main computational bottleneck of EDNN: each step requires the solution of a dense linear least-squares problem in the full parameter space. As the number of trainable parameters grows, this dense solve quickly dominates the cost.

Recent work has therefore explored reduced-update strategies for
sequential-in-time neural solvers. Randomized sparse neural Galerkin
schemes reduce cost by updating sparse random subsets of parameters~\cite{berman2023randomized}. CoLoRA exploits continuously varying low-rank structures for reduced implicit neural modeling of parameterized PDEs~\cite{berman2024colora}. pETNNs combine evolutionary neural networks with tensorized parameterizations and partial parameter updates~\cite{kao2024petnns}. These developments make clear that the design of efficient, dynamically meaningful reduced trial spaces is now a central issue in sequential neural PDE solvers. A closely related non-neural line of work evolves the parameters of a nonlinear reduced parametrization by an analogous residual-minimization principle: shape-morphing reduced-order models advance a low-dimensional solution ansatz in time, and have been developed with conservation and information-geometric structure~\cite{anderson2022evolution, anderson2022shape, anderson2024fisher}.

The present work addresses this issue from a geometric low-rank
viewpoint. A direct low-rank factorization of the EDNN velocity reduces the number of unknowns, but it destroys the linear structure of the EDNN least-squares problem and leads to a bilinear optimization at every time step. Our key idea is to retain the low-rank intuition while preserving the linear least-squares structure: instead of evolving low-rank factors directly, we construct an adaptive tangent space from a truncated singular value decomposition (SVD) of the current layer weights and solve the EDNN problem projected onto that tangent space.

This interpretation is important. LR-EDNN is not merely a heuristic
low-rank constraint or a static parameter-efficient update. Rather, it
is a projected sequential-in-time neural solver whose admissible update space is determined by the current state of the network. The
tangent-space formulation also gives the method a precise geometric link to dynamical low-rank approximation~\cite{koch2007dynamical, lubich2014projector}. At the same time,
unlike classical rank-constrained integrators, our method acts on the
parameter velocity only. It does not enforce that the discrete
parameter trajectory remain on a fixed-rank manifold. This distinction
is conceptually and practically important.

The main contributions of this work are as follows:
\begin{enumerate}
\def\labelenumi{\arabic{enumi}.}
\item
  \textbf{Projected reduced EDNN formulation.} We recast low-rank EDNN
  as a projected least-squares evolution problem on an adaptive reduced trial space determined by the current layer weights. Using the global tangent-space parameterization matrix \(B\), we construct the reduced Jacobian \(J_\theta B\) directly through layerwise Jacobian--vector products, without forming the full Jacobian \(J_\theta\).
\item
  \textbf{Geometric characterization of the reduced space.} We prove
  that the admissible layerwise update space is exactly the tangent
  space of the fixed-rank matrix manifold at the truncated layer
  weights, making the geometric content of the reduction explicit.
\item
  \textbf{Approximation results.} We show that the
  reduced update is the best approximation of the full EDNN update in a Jacobian-induced seminorm, derive a residual decomposition identity, and establish uniqueness modulo the Jacobian nullspace. We also establish a finite-time comparison estimate showing that the LR-EDNN trajectory differs from the full EDNN trajectory by the accumulated tangent-space projection defects, with amplification governed by the assumed Lipschitz and directional-coercivity constants.
\item
  \textbf{Numerical evidence across representative nonlinear PDEs.} We
  demonstrate on nonlinear diffusion, phase-field, and
  convection-diffusion benchmarks that the proposed reduction can
  significantly decrease the computational cost of EDNN while retaining the solution quality of the full model when the rank is sufficiently large.
\end{enumerate}

The remainder of the paper is organized as follows. Section 2 reviews
EDNN and makes explicit its projection structure. Section 3 develops the adaptive tangent-space reduction, establishes its approximation
properties, and derives the reduced linear system. Section 4 reports
numerical experiments. Section 5 discusses accuracy, efficiency, and the relation of LR-EDNN to other reduced neural solvers. Section 6
concludes.

%%%%%%%%%%%%%%%%%%%%%%%%%%%%%%%%%%%%%%%%%%%%%%%%%%%%%%%%%%%%%%%%%%%%
%%%%%   Section 2
%%%%%%%%%%%%%%%%%%%%%%%%%%%%%%%%%%%%%%%%%%%%%%%%%%%%%%%%%%%%%%%%%%%%
\section{EDNN as a sequential projected neural solver}

\subsection{Problem setting and EDNN
formulation}\label{problem-setting-and-ednn-formulation}

We consider a time-dependent PDE of the form \begin{equation}\label{eq:e1}
\partial_t u = \mathcal{N}_x(u), \qquad x \in \Omega \subset \mathbb{R}^d,\quad t \in [0,T],
\end{equation} where \(u(x,t)\) denotes the solution and \(\mathcal{N}_x\) is a
spatial differential operator.

In the EDNN framework, the spatial solution at time \(t\) is represented
by a neural network \[
\widehat{u}(x; w(t)),
\] with time-dependent parameter vector \(w(t)\in \mathbb{R}^P\). We
distinguish this global vectorized parameter from the layerwise weight
matrices \[
W_\ell(t)\in \mathbb{R}^{n_\ell \times m_\ell}, \qquad \ell=1,\dots,L.
\] Absorbing biases into augmented matrices for notational simplicity,
we write \begin{equation}\label{eq:e2}
w(t) =
\begin{bmatrix}
\mathrm{vec}(W_1(t))\\
\vdots\\
\mathrm{vec}(W_L(t))
\end{bmatrix},
\qquad
P = \sum_{\ell=1}^L n_\ell m_\ell.
\end{equation}

The initial parameter state is obtained by fitting the initial
condition: \begin{equation}\label{eq:e3}
\mathcal{L}_0(w(0)) =
\frac12 \sum_{i=1}^{M_0}
\left\|
\widehat{u}(x_i;w(0)) - u(x_i,0)
\right\|_2^2,
\end{equation} where \(\{x_i\}_{i=1}^{M_0}\subset \Omega\) is a set of training
points.

After initialization, the PDE evolution is transferred to parameter
space. By the chain rule, \begin{equation}\label{eq:e4}
\partial_t \widehat{u}(x;w(t))
=
\frac{\partial \widehat{u}}{\partial w}(x;w(t))\,\dot{w}(t),
\end{equation} where \(\dot{w}(t)\in \mathbb{R}^P\) is the parameter velocity. At
time \(t_n=n\Delta t\), we write \(w^n:=w(t_n)\) and define collocation
points \(\{x_i\}_{i=1}^{M_c}\). Stacking all scalar outputs, we define
the Jacobian and PDE vector \begin{equation}\label{eq:e5}
J^n :=
\begin{bmatrix}
\dfrac{\partial \widehat{u}}{\partial w}(x_1;w^n)\\
\vdots\\
\dfrac{\partial \widehat{u}}{\partial w}(x_{M_c};w^n)
\end{bmatrix}
\in \mathbb{R}^{M\times P},
\qquad
N^n :=
\begin{bmatrix}
\mathcal{N}_x(\widehat{u}(x_1;w^n))\\
\vdots\\
\mathcal{N}_x(\widehat{u}(x_{M_c};w^n))
\end{bmatrix}
\in \mathbb{R}^{M}.
\end{equation}

The EDNN update is then defined by the least-squares problem \begin{equation}\label{eq:e6}
\dot{w}_{\mathrm{opt}}^n
\in
\operatorname*{argmin}_{\dot{w}\in \mathbb{R}^P}
\frac12 \|J^n \dot{w} - N^n\|_2^2.
\end{equation} Its normal equations are \begin{equation}\label{eq:e7}
(J^n)^\top J^n \dot{w}_{\mathrm{opt}}^n
=
(J^n)^\top N^n.
\end{equation} A time integrator, for example explicit Euler, \begin{equation}\label{eq:e8}
w^{n+1}=w^n+\Delta t\,\dot{w}_{\mathrm{opt}}^n,
\end{equation} then advances the neural solution.

\subsection{Projection viewpoint and computational
bottleneck}\label{projection-viewpoint-and-computational-bottleneck}

The least-squares problem \eqref{eq:e6} has a clear projection interpretation. The
quantity \(J^n \dot{w}\) is the discrete solution derivative induced at
the collocation points by a parameter velocity \(\dot{w}\). Thus EDNN
chooses the parameter velocity whose induced discrete derivative best
matches the PDE right-hand side \(N^n\).

This viewpoint is useful for two reasons. First, it makes precise what
EDNN is computing at each time step: a local projection of PDE dynamics
into the tangent space of the current network representation. Second, it
clarifies the source of the computational bottleneck: the dense solve in
\eqref{eq:e7} is performed in the full parameter space \(\mathbb{R}^P\).

For modern neural architectures, \(P\) may be large even when the number
of time steps and collocation points is moderate. The per-step cost of
solving a dense \(P\times P\) system can therefore dominate the overall
runtime. Our objective is to reduce this cost without changing the
sequential-in-time character of EDNN and without replacing its linear
least-squares structure by a nonlinear inner optimization.

%%%%%%%%%%%%%%%%%%%%%%%%%%%%%%%%%%%%%%%%%%%%%%%%%%%%%%%%%%%%%%%%%%%%
%%%%%   Section 3
%%%%%%%%%%%%%%%%%%%%%%%%%%%%%%%%%%%%%%%%%%%%%%%%%%%%%%%%%%%%%%%%%%%%

\section{Adaptive tangent-space reduction}

\subsection{From low-rank velocity ansatz to tangent-space
projection}\label{from-low-rank-velocity-ansatz-to-tangent-space-projection}

Let \(r\in\mathbb{N}\) be a global rank cap. For each
bias-augmented layer matrix
\(W_\ell^n\in\mathbb{R}^{n_\ell\times m_\ell}\), define the effective
layer rank
\begin{equation}\label{eq:layer-rank}
r_\ell=r_\ell(r):=\min\{r,n_\ell,m_\ell\}.
\end{equation}
For an affine layer with input and output widths \(d_{\ell-1}\) and
\(d_\ell\), respectively, \(n_\ell=d_\ell\) and
\(m_\ell=d_{\ell-1}+1\), where the extra column contains the bias.
Thus the cap \(r\) is common to all layers, while the admissible
truncation ranks \(r_\ell\) may differ because the layer matrix
dimensions differ.

A natural attempt to reduce the number of unknowns is to impose a
low-rank structure on the velocity of each layer weight matrix: 
\begin{equation}\label{eq:e9}{
    \dot{W}_\ell = A_\ell B_\ell,
    \qquad
    A_\ell \in \mathbb{R}^{n_\ell\times r_\ell},\quad
    B_\ell \in \mathbb{R}^{r_\ell\times m_\ell}.}
\end{equation} 
The corresponding factor-coordinate count in layer \(\ell\) is
\(r_\ell(n_\ell+m_\ell)\), instead of \(n_\ell m_\ell\) full
parameters.

However, substituting \eqref{eq:e9} into the EDNN objective destroys linearity. If
\(J_\ell^n\) denotes the block of \(J^n\) corresponding to \(W_\ell^n\),
then one obtains the bilinear problem \begin{equation}\label{eq:e10}
\min_{\{A_\ell,B_\ell\}_{\ell=1}^L}
\frac12
\left\|
\sum_{\ell=1}^L J_\ell^n \,\mathrm{vec}(A_\ell B_\ell) - N^n
\right\|_2^2,
\end{equation} which is nonconvex in the factors \(\{A_\ell,B_\ell\}\). Solving such
a nonlinear inner problem at every time step would undermine the
computational advantage of the reduction.

We therefore seek a linear reduced model that captures the
dominant low-rank update directions without directly evolving low-rank
factors.

Fix a time level \(t_n\) and let 
\begin{equation}\label{eq:e11}
{
W_\ell^n \approx W_{\ell,r_\ell}^n
=U_{\ell,r_\ell}\Sigma_{\ell,r_\ell}V_{\ell,r_\ell}^\top
}
\end{equation} 
be the rank-\(r_\ell\) truncated SVD of the current layer matrix, with
\[
{
U_{\ell,r_\ell}\in \mathbb{R}^{n_\ell\times r_\ell},\qquad
V_{\ell,r_\ell}\in \mathbb{R}^{m_\ell\times r_\ell},\qquad
\Sigma_{\ell,r_\ell}\in \mathbb{R}^{r_\ell\times r_\ell},
}
\] and strictly positive retained singular values.

For each layer we define the fixed-rank manifold \begin{equation}\label{eq:e12}
{
\mathcal{M}_{r_\ell,\ell} := \left\{X\in \mathbb{R}^{n_\ell\times m_\ell} : \mathrm{rank}(X)=r_\ell\right\}.
}
\end{equation} 
Its tangent space at \(X\in \mathcal{M}_{r_\ell,\ell}\) is \begin{equation}\label{eq:e13}
{
T_X\mathcal{M}_{r_\ell,\ell}
:=
\left\{
\dot{\Gamma}(0) :
\Gamma:(-\varepsilon,\varepsilon)\to \mathcal{M}_{r_\ell,\ell}
\text{ is } C^1,\ \Gamma(0)=X
\right\}.
}
\end{equation}

Let
\(U_{\ell,r_\ell,\perp}\) and \(V_{\ell,r_\ell,\perp}\) be orthonormal
complements of \(U_{\ell,r_\ell}\) and \(V_{\ell,r_\ell}\). Then \begin{equation}\label{eq:e14}
{
T_{W_{\ell,r_\ell}^n}\mathcal{M}_{r_\ell,\ell}
=
\left\{
U_{\ell,r_\ell}S_\ell V_{\ell,r_\ell}^\top
+
U_{\ell,r_\ell,\perp}P_\ell V_{\ell,r_\ell}^\top
+
U_{\ell,r_\ell}Q_\ell V_{\ell,r_\ell,\perp}^\top
\right\},
}
\end{equation} where \[
{
S_\ell \in \mathbb{R}^{r_\ell\times r_\ell},\quad
P_\ell \in \mathbb{R}^{(n_\ell-r_\ell)\times r_\ell},\quad
Q_\ell \in \mathbb{R}^{r_\ell\times (m_\ell-r_\ell)}.
}
\] Equivalently, \begin{equation}\label{eq:e15}
{
T_{W_{\ell,r_\ell}^n}\mathcal{M}_{r_\ell,\ell}
=
\left\{
L_\ell V_{\ell,r_\ell}^\top + U_{\ell,r_\ell} K_\ell^\top :
L_\ell \in \mathbb{R}^{n_\ell\times r_\ell},\
K_\ell \in \mathbb{R}^{m_\ell\times r_\ell}
\right\}.
}
\end{equation} Hence \begin{equation}\label{eq:e16}
{
\dim T_{W_{\ell,r_\ell}^n}\mathcal{M}_{r_\ell,\ell}
=r_\ell(n_\ell + m_\ell - r_\ell).
}
\end{equation}

Now define balanced factors \begin{equation}\label{eq:e17}
{
X_\ell^n := U_{\ell,r_\ell}\Sigma_{\ell,r_\ell}^{1/2}\in \mathbb{R}^{n_\ell\times r_\ell},
\qquad
Y_\ell^n := \Sigma_{\ell,r_\ell}^{1/2}V_{\ell,r_\ell}^\top \in \mathbb{R}^{r_\ell\times m_\ell}.
}
\end{equation} 
Thus \(W_{\ell,r_\ell}^n = X_\ell^n Y_\ell^n\). A first-order
perturbation of these factors produces \begin{equation}\label{eq:e18}
\delta W_\ell = \delta X_\ell\,Y_\ell^n + X_\ell^n\,\delta Y_\ell.
\end{equation} Motivated by \eqref{eq:e18}, we define the admissible reduced update space \begin{equation}\label{eq:e19}
{
\mathcal{T}_{r_\ell}(W_\ell^n)
:=
\left\{
C_\ell Y_\ell^n + X_\ell^n D_\ell :
C_\ell \in \mathbb{R}^{n_\ell\times r_\ell},\
D_\ell \in \mathbb{R}^{r_\ell\times m_\ell}
\right\}.
}
\end{equation}

\paragraph{Lemma 3.1}
Assume that the retained singular values in \(\Sigma_{\ell,r_\ell}\) are
positive. Then \begin{equation}\label{eq:e20}
{
\mathcal{T}_{r_\ell}(W_\ell^n)
=T_{W_{\ell,r_\ell}^n}\mathcal{M}_{r_\ell,\ell}.
}
\end{equation}

\paragraph{Proof}
Using \eqref{eq:e17}, \[
{
C_\ell Y_\ell^n + X_\ell^n D_\ell
=
(C_\ell \Sigma_{\ell,r_\ell}^{1/2})V_{\ell,r_\ell}^\top
+
U_{\ell,r_\ell}(D_\ell^\top \Sigma_{\ell,r_\ell}^{1/2})^\top.
}
\]
Thus every element of \(\mathcal{T}_{r_\ell}(W_\ell^n)\) has the form \eqref{eq:e15}.
Conversely, given \(L_\ell\) and \(K_\ell\) in \eqref{eq:e15}, choose \[
{
C_\ell = L_\ell \Sigma_{\ell,r_\ell}^{-1/2},
\qquad
D_\ell = \Sigma_{\ell,r_\ell}^{-1/2}K_\ell^\top.
}
\] Then \[
{
C_\ell Y_\ell^n + X_\ell^n D_\ell
=L_\ell V_{\ell,r_\ell}^\top + U_{\ell,r_\ell}K_\ell^\top.
}
\] Thus the two spaces coincide. \(\square\)

The lemma shows that our reduced model is not an ad hoc linearization:
it is exactly the tangent space of the fixed-rank matrix manifold at the
truncated layer state.

\paragraph{Remark 3.1}
The space \(\mathcal{T}_{r_\ell}(W_\ell^n)\) depends only on the retained left
and right singular \emph{subspaces}, not on the particular orthonormal
bases chosen to represent them. Hence the reduced trial space is
invariant under orthogonal changes of basis in the rank-\(r_\ell\) singular
subspaces.

\paragraph{Remark 3.2}
{
When \(r_\ell<\min\{n_\ell,m_\ell\}\), a strict gap
\(\sigma_{\ell,r_\ell}>\sigma_{\ell,r_\ell+1}\) uniquely determines the
retained singular subspaces and hence the tangent space. If the two
singular values are equal, the rank-\(r_\ell\) truncation may be
nonunique.
}

\paragraph{Remark 3.3}
LR-EDNN restricts the \emph{parameter velocity} to
\(\mathcal{T}_{r_\ell}(W_\ell^n)\) but does not enforce that \(W_\ell^{n+1}\)
itself remain on a rank-\(r_\ell\) manifold. The method is therefore a
first-order tangent-space approximation of low-rank dynamics, not a
rank-constrained time integrator. Throughout, the global rank cap \(r\) is fixed during a simulation; the term \emph{adaptive} refers only to the reduced tangent space being recomputed from the evolving weights at every time step.

\subsection{Reduced formulation and algorithm}
\label{reduced-linear-system-algorithm-and-implementation}

Vectorizing each admissible layerwise update and stacking all layers, we
define the global reduced space \begin{equation}\label{eq:e21}
{
\mathcal{T}_r(w^n)
:=
\left\{
\begin{bmatrix}
\mathrm{vec}(\Delta W_1)\\
\vdots\\
\mathrm{vec}(\Delta W_L)
\end{bmatrix}
:
\Delta W_\ell \in \mathcal{T}_{r_\ell}(W_\ell^n),\ \ell=1,\dots,L
\right\}
\subset \mathbb{R}^P.
}
\end{equation}

The reduced LR-EDNN update is then \begin{equation}\label{eq:e22}
\dot{w}_r^n
\in
\operatorname*{argmin}_{\zeta\in \mathcal{T}_r(w^n)}
\frac12 \|J^n \zeta - N^n\|_2^2.
\end{equation} Thus LR-EDNN solves \emph{the same} local EDNN least-squares problem
as the full method, but on a smaller, adaptive trial space.

{
For each layer, define
\begin{equation}\label{eq:e41}
q_\ell:=r_\ell(n_\ell+m_\ell),
\qquad
\gamma_\ell :=
\begin{bmatrix}
\mathrm{vec}(C_\ell)\\
\mathrm{vec}(D_\ell)
\end{bmatrix}
\in \mathbb{R}^{q_\ell}.
\end{equation}
Vectorizing the layerwise update
\(\dot{W}_\ell=C_\ell Y_\ell^n+X_\ell^nD_\ell\),
we define the layerwise tangent-space parameterization matrix
\begin{equation}\label{eq:e42}
\Phi_\ell^n
=
\begin{bmatrix}
(Y_\ell^n)^\top\otimes I_{n_\ell}
\quad
I_{m_\ell}\otimes X_\ell^n
\end{bmatrix}
\in
\mathbb{R}^{n_\ell m_\ell \times q_\ell}.
\end{equation}
Then
\begin{equation}\label{eq:e43}
\mathrm{vec}(\dot{W}_\ell)=\Phi_\ell^n \gamma_\ell.
\end{equation}

Stacking the layerwise coordinates gives
\begin{equation}\label{eq:e44}
\gamma :=
\begin{bmatrix}
\gamma_1\\
\vdots\\
\gamma_L
\end{bmatrix}
\in \mathbb{R}^{q_r},
\qquad
q_r := \sum_{\ell=1}^L q_\ell
=\sum_{\ell=1}^L r_\ell(n_\ell+m_\ell),
\end{equation}
and define the global tangent-space parameterization matrix
\begin{equation}\label{eq:e45}
B^n := \mathrm{blkdiag}(\Phi_1^n,\dots,\Phi_L^n).
\end{equation}
Then
\begin{equation}\label{eq:e46}
\dot{w}=B^n\gamma.
\end{equation}

By Lemma~3.1 and the block-diagonal construction,
\begin{equation}\label{eq:range-B}
\operatorname{range}(B^n)=\mathcal{T}_r(w^n).
\end{equation}

Define the reduced Jacobian matrix \(G^n:=J^nB^n\).
Substituting \eqref{eq:e46} into \eqref{eq:e22} yields
\begin{equation}\label{eq:e47}
\gamma_{\mathrm{opt}}^n
\in
\operatorname*{argmin}_{\gamma}
\frac12 \|G^n\gamma - N^n\|_2^2.
\end{equation}
The corresponding normal equations are
\begin{equation}\label{eq:e48}
(G^n)^\top G^n\gamma=(G^n)^\top N^n.
\end{equation}
The full parameter velocity is reconstructed by
\begin{equation}\label{eq:e49}
\dot{w}_r^n = B^n\gamma_{\mathrm{opt}}^n.
\end{equation}
}

\begin{algorithm}
\caption{Low-Rank Evolutionary Deep Neural Network (LR-EDNN)}
\label{alg:lrednn}
\begin{algorithmic}[1]
\State {\textbf{Input:} Initial condition \(u(x,0)\), PDE operator \(\mathcal{N}_x\), global rank cap \(r\), time step \(\Delta t\), final time \(T\).}
\State Fit the initial condition and obtain the initial parameter state \(w^0\).
\For{\(n = 0,1,\dots,(T/\Delta t)-1\)}
    \ForAll{layers \(\ell=1,\dots,L\)}
        \State {Set \(r_\ell=\min\{r,n_\ell,m_\ell\}\).}
        \State {Compute the rank-\(r_\ell\) truncated SVD \(W_\ell^n \approx U_{\ell,r_\ell}\Sigma_{\ell,r_\ell}V_{\ell,r_\ell}^\top\).}
        \State {Form \(X_\ell^n = U_{\ell,r_\ell}\Sigma_{\ell,r_\ell}^{1/2}\) and \(Y_\ell^n = \Sigma_{\ell,r_\ell}^{1/2}V_{\ell,r_\ell}^\top\).}
        \State {Build the layerwise tangent-space parameterization matrix \(\Phi_\ell^n\) from Eq.~\ref{eq:e42}.}
    \EndFor
    \State {Assemble \(B^n\) from Eq.~\ref{eq:e45} and evaluate \(N^n\) at the collocation points.}
    \State {Construct the global reduced Jacobian matrix \(G^n=J^nB^n\) directly by layerwise Jacobian--vector products; do not form the full Jacobian \(J^n\).}
    \State {Using a fixed deterministic least-squares convention, solve problem \eqref{eq:e47} for the selected \(\gamma_{\mathrm{opt}}^n\).}
    \State {Recover the full parameter velocity \(\dot{w}_r^n = B^n\gamma_{\mathrm{opt}}^n\).}
    \State Update the parameters, e.g.,
    \[
    w^{n+1} = w^n + \Delta t\,\dot{w}_r^n.
    \]
\EndFor
\State \textbf{Output:} The solution trajectory \(\widehat{u}(x,t)\).
\end{algorithmic}
\end{algorithm}

\subsection{Approximation properties}
\label{projected-reduced-ednn-and-approximation-properties}

The optimality condition for the reduced problem \eqref{eq:e22} is
\begin{equation}\label{eq:e23}
\langle J^n\eta,\, J^n \dot{w}_r^n - N^n\rangle = 0,
\qquad
\forall\,\eta\in \mathcal{T}_r(w^n).
\end{equation}

Since the objective depends on \(\zeta\) only through \(J^n\zeta\), the
natural metric is the Jacobian-induced seminorm
\begin{equation}\label{eq:e24}
\|z\|_{J^n} := \|J^n z\|_2,
\qquad z\in \mathbb{R}^P.
\end{equation}
If \(J^n\) has a nontrivial nullspace, this is only a seminorm:
different parameter velocities may induce the same discrete solution
derivative.

\paragraph{Theorem 3.1 (Best approximation in the Jacobian seminorm)}
Let \(\dot{w}_{\mathrm{opt}}^n\) be any minimizer of the full EDNN
problem \eqref{eq:e6}, and let \(\dot{w}_r^n\) be any minimizer of the reduced problem \eqref{eq:e22}.
Then
\begin{equation}\label{eq:e25}
\dot{w}_r^n
\in
\operatorname*{argmin}_{\zeta\in \mathcal{T}_r(w^n)}
\|\zeta - \dot{w}_{\mathrm{opt}}^n\|_{J^n}^2.
\end{equation}
Equivalently,
\begin{equation}\label{eq:e26}
\dot{w}_r^n
\in
\operatorname*{argmin}_{\zeta\in \mathcal{T}_r(w^n)}
\|J^n(\zeta-\dot{w}_{\mathrm{opt}}^n)\|_2^2.
\end{equation}
Moreover,
\begin{equation}\label{eq:e27}
\|J^n\dot{w}_r^n - N^n\|_2^2
=
\|J^n(\dot{w}_r^n-\dot{w}_{\mathrm{opt}}^n)\|_2^2
+
\|J^n\dot{w}_{\mathrm{opt}}^n - N^n\|_2^2.
\end{equation}

\paragraph{Proof}
Define the full residual
\[
r_{\mathrm{opt}}^n := J^n\dot{w}_{\mathrm{opt}}^n - N^n.
\]
Since \(\dot{w}_{\mathrm{opt}}^n\) minimizes \eqref{eq:e6},
\[
(J^n)^\top r_{\mathrm{opt}}^n = 0.
\]
Hence, for any \(\zeta\in \mathbb{R}^P\),
\[
\|J^n\zeta-N^n\|_2^2
=
\|J^n(\zeta-\dot{w}_{\mathrm{opt}}^n)+r_{\mathrm{opt}}^n\|_2^2
=
\|J^n(\zeta-\dot{w}_{\mathrm{opt}}^n)\|_2^2
+
\|r_{\mathrm{opt}}^n\|_2^2,
\]
because the cross term vanishes:
\[
\langle J^n(\zeta-\dot{w}_{\mathrm{opt}}^n),r_{\mathrm{opt}}^n\rangle
=
\langle \zeta-\dot{w}_{\mathrm{opt}}^n,(J^n)^\top r_{\mathrm{opt}}^n\rangle
=
0.
\]
Restricting \(\zeta\) to \(\mathcal{T}_r(w^n)\) yields
\eqref{eq:e25}--\eqref{eq:e26}, and evaluating at
\(\zeta=\dot{w}_r^n\) gives \eqref{eq:e27}. \(\square\)

Theorem 3.1 identifies LR-EDNN as a best approximation in the Jacobian-induced seminorm of the full EDNN update onto the adaptive tangent space, measured in the quantity that matters for the PDE residual: the discrete solution derivative \(J^n\dot{w}\).

\paragraph{Lemma 3.2}
If \(\dot{w}_{r,1}^n,\dot{w}_{r,2}^n\in \mathcal{T}_r(w^n)\) both
minimize \eqref{eq:e22}, then
\begin{equation}\label{eq:e28}
J^n \dot{w}_{r,1}^n = J^n \dot{w}_{r,2}^n.
\end{equation}
Equivalently,
\begin{equation}\label{eq:e29}
\dot{w}_{r,1}^n - \dot{w}_{r,2}^n \in \ker(J^n)\cap \mathcal{T}_r(w^n).
\end{equation}
In particular, the reduced minimizer is unique whenever
\begin{equation}\label{eq:e30}
\ker(J^n)\cap \mathcal{T}_r(w^n)=\{0\}.
\end{equation}

\paragraph{Proof}
The reduced objective is strictly convex in the image variable
\(J^n\zeta\). If two minimizers had different images under \(J^n\),
their midpoint would yield a strictly smaller objective value,
contradicting minimality. The rest follows immediately. \(\square\)

Lemma 3.2 clarifies an important point: even when the parameter velocity is not unique, the induced discrete solution derivative is unique.

To connect the reduced and full updates at the solution level, define
the stacked network output at the collocation points:
\begin{equation}\label{eq:e31}
\widehat{U}(w) :=
\begin{bmatrix}
\widehat{u}(x_1;w)\\
\vdots\\
\widehat{u}(x_{M_c};w)
\end{bmatrix},
\qquad
D\widehat{U}(w^n)=J^n.
\end{equation}

\paragraph{{Theorem 3.2 (Finite-time comparison with full EDNN)}}
{
Write \(J(w):=D\widehat{U}(w)\), the EDNN Jacobian as a function of the parameters (cf.\ \eqref{eq:e31}), and let \(N(w)\) be the PDE operator vector evaluated at \(w\). Fix deterministic conventions for the full least-squares solve and the reduced coefficient least-squares solve in Algorithm~\ref{alg:lrednn}. Define the selected vector fields
\begin{equation}\label{eq:Ffield}
F(w)\in\operatorname*{argmin}_{v\in\mathbb{R}^P}\tfrac12\|J(w)v-N(w)\|_2^2,
\qquad
F_r(w)\in\operatorname*{argmin}_{v\in\mathcal{T}_r(w)}\tfrac12\|J(w)v-N(w)\|_2^2.
\end{equation}
Here \(F_r(w)\) is the parameter velocity returned by Algorithm~\ref{alg:lrednn} under these conventions. The fixed conventions make both fields single valued; moreover, Lemma~3.2 shows that \(J(w)F_r(w)\) is independent of the selected reduced minimizer.} Over a horizon \(N\), the full and reduced explicit-Euler trajectories
\begin{equation}\label{eq:traj}
w^{m+1}=w^m+\Delta t\,F(w^m),\qquad w_r^{m+1}=w_r^m+\Delta t\,F_r(w_r^m),\qquad m=0,\dots,N-1,
\end{equation}
coincide with full EDNN and Algorithm~\ref{alg:lrednn}, respectively. Let the local reduction defect be
\begin{equation}\label{eq:defect}
\eta_r(w):=\inf_{\zeta\in\mathcal{T}_r(w)}\|J(w)(\zeta-F(w))\|_2=\|J(w)(F_r(w)-F(w))\|_2,
\end{equation}
the second equality by Theorem 3.1. Assume that, on a compact set \(\mathcal{K}\) containing both trajectories for \(0\le m\le N\):
\begin{itemize}
\item[\textup{(A1)}] \(F\) is Lipschitz on \(\mathcal{K}\): \(\|F(v)-F(z)\|_2\le L_F\|v-z\|_2\);
\item[\textup{(A2)}] \(J(w)\) is uniformly coercive along the reduction direction \(d(w):=F_r(w)-F(w)\): \(\|d(w)\|_2\le\alpha^{-1}\|J(w)\,d(w)\|_2\) for some \(\alpha>0\).
\end{itemize}
Then
\begin{equation}\label{eq:gronwall}
\|w_r^m-w^m\|_2\le(1+\Delta t\,L_F)^m\|w_r^0-w^0\|_2+\frac{\Delta t}{\alpha}\sum_{k=0}^{m-1}(1+\Delta t\,L_F)^{\,m-1-k}\,\eta_r(w_r^k),\qquad 0\le m\le N.
\end{equation}
In particular, if \(w_r^0=w^0\) and \(\eta_r(w_r^k)\le\bar\eta_r\) for all \(k\), then for \(0\le m\le N\),
\begin{equation}\label{eq:gronwallunif}
\|w_r^m-w^m\|_2
\le \frac{\bar\eta_r}{\alpha}
\begin{cases}
\dfrac{(1+\Delta t\,L_F)^m-1}{L_F}, & L_F>0,\\[1.2ex]
m\Delta t, & L_F=0.
\end{cases}
\end{equation}
If, in addition, \(\widehat{U}\) is Lipschitz on \(\mathcal{K}\) with constant \(L_U\), the solution-level error obeys
\begin{equation}\label{eq:outputbound}
\|\widehat{U}(w_r^m)-\widehat{U}(w^m)\|_2\le L_U\,\|w_r^m-w^m\|_2,\qquad 0\le m\le N.
\end{equation}

\paragraph{{Proof}}
{
Set \(e^m:=w_r^m-w^m\). Subtracting the two updates in \eqref{eq:traj} and adding and subtracting \(F(w_r^m)\) gives
\begin{equation}\label{eq:proofsplit}
e^{m+1}=e^m+\Delta t\big[F(w_r^m)-F(w^m)\big]+\Delta t\big[F_r(w_r^m)-F(w_r^m)\big].
\end{equation}
By (A1), \(\|F(w_r^m)-F(w^m)\|_2\le L_F\|e^m\|_2\); by (A2) and \eqref{eq:defect}, \(\|F_r(w_r^m)-F(w_r^m)\|_2\le\alpha^{-1}\eta_r(w_r^m)\). Hence
\begin{equation}\label{eq:onestepbound}
\|e^{m+1}\|_2\le(1+\Delta t\,L_F)\|e^m\|_2+\frac{\Delta t}{\alpha}\,\eta_r(w_r^m),
\end{equation}
and the discrete Gr\"onwall inequality gives \eqref{eq:gronwall}. The special case \eqref{eq:gronwallunif} and the solution-level bound \eqref{eq:outputbound} are immediate. \(\square\)
}

\paragraph{Remark 3.4 (Limitation of the layerwise trial space)}
The layerwise SVD constructs \(\mathcal{T}_r(w^n)\), but it does not guarantee that the projection defect \(\eta_r\) in \eqref{eq:defect} is small; this defect is the relevant measure of accuracy relative to full EDNN.

\subsection{Complexity and implementation
remarks}\label{complexity-and-implementation-remarks}

Let \[
P=\sum_{\ell=1}^L n_\ell m_\ell
\] be the full parameter dimension. {Full EDNN uses the
full Jacobian \(J^n\in\mathbb{R}^{M\times P}\). LR-EDNN instead uses the
directly constructed reduced Jacobian matrix
\(G^n=J^nB^n\in\mathbb{R}^{M\times q_r}\), where} \[
{
q_r=\sum_{\ell=1}^L r_\ell(n_\ell+m_\ell).
}
\]

The additional per-step overhead comes from the truncated SVDs of the
layer matrices. For full SVDs, the cost is bounded by \[
\sum_{\ell=1}^L O\!\left(n_\ell m_\ell \min(n_\ell,m_\ell)\right),
\] whereas {truncated rank-\(r_\ell\) decompositions} typically scale more
favorably, approximately as \[
{
\sum_{\ell=1}^L O(n_\ell m_\ell r_\ell),
}
\] up to orthogonalization and iteration costs.

{Constructing \(G^n=J^nB^n\) directly avoids storing
the full \(M\times P\) Jacobian and requires only \(O(Mq_r)\)
reduced-Jacobian storage. If a dense reduced normal matrix is assembled,}
assembling the reduced normal matrix costs \(O(Mq_r^2)\) and solving it
costs \(O(q_r^3)\), compared with \(O(MP^2)\) and \(O(P^3)\) for the
explicit full EDNN solve. Thus the central algebraic benefit of LR-EDNN
is the replacement of {\(P\) parameter columns by
\(q_r\) reduced coordinate columns}, with \(q_r\ll P\) in the low-rank
regime of interest.

%%%%%%%%%%%%%%%%%%%%%%%%%%%%%%%%%%%%%%%%%%%%%%%%%%%%%%%%%%%%%%%%%%%%%%
%%% Section 4
%%%%%%%%%%%%%%%%%%%%%%%%%%%%%%%%%%%%%%%%%%%%%%%%%%%%%%%%%%%%%%%%%%%%%%
\section{Numerical Experiments}\label{sec:numerical}

This section evaluates the proposed LR-EDNN method through a sequence of benchmark PDE problems with increasing problem complexity and network scale. For each example, we first describe the PDE setting, including the domain, boundary conditions, initial condition, and reference solver. We then specify the neural network architecture and the EDNN update configuration used for the comparison. The results are presented using final-solution and pointwise-error plots, together with timing measurements that separate the Jacobian construction, SVD or basis construction, linear solve, and remaining computational cost. This organization allows each experiment to be read independently, while also showing how the computational behavior changes from smaller scalar problems to larger two-dimensional systems. Section~5 then discusses the trends across all examples, focusing on accuracy, runtime reduction, the role of the low-rank tangent space, and the effect of increasing model size.

All experiments were implemented in Python 3.12.13 using PyTorch 2.3.1 with CUDA 12.1. The computations were performed on NVIDIA A100 80GB PCIe GPUs. Unless otherwise stated, each experiment used one GPU through PyTorch's default CUDA device. All neural network parameters, PDE residual evaluations, Jacobian or reduced-Jacobian matrices, SVD bases, and linear solves were computed in double precision.

On this GPU platform, dense and iterative linear algebra operations are highly optimized. As a result, the linear solve time is often smaller than the time required to construct the Jacobian or reduced Jacobian matrix. If the same experiments were carried out on CPU, the relative advantage of LR-EDNN in the linear algebra stage would likely be larger, since LR-EDNN replaces the full least-squares problem with a much smaller reduced problem. However, CPU computation becomes difficult for the larger dense EDNN systems as the number of parameters increases. We therefore report GPU results to provide a practical and conservative comparison in a setting where full EDNN is still computationally feasible and where the measured speedup is not simply caused by slow CPU linear algebra.

For each PDE example, we compare full EDNN, randomized sparse EDNN, and the corresponding low-rank EDNN variant. Full EDNN forms the full parameter Jacobian \(J_\theta\). Randomized sparse EDNN forms only the selected active columns of \(J_\theta\). LR-EDNN forms the reduced Jacobian matrix \(J_\theta B\) directly by tangent propagation, where \(B\) denotes the global tangent-space parameterization matrix. Thus, each method is timed according to the matrix it actually constructs.

Accuracy is reported using the final relative \(L^2\) error,
\[
    \frac{\|u_\theta(\cdot,T)-u_{\mathrm{ref}}(\cdot,T)\|_2}
    {\|u_{\mathrm{ref}}(\cdot,T)\|_2},
\]
where \(u_{\mathrm{ref}}\) is the trusted reference solution. For the two-dimensional Burgers system, this norm is taken over both velocity components. Runtime is decomposed into Jacobian construction, SVD or basis construction, linear solve, and all remaining operations. This timing decomposition is used both in the individual examples and in the cross-example discussion in Section~5.

\subsection{Example 1: Porous Medium Equation}

We first consider a two-dimensional porous medium equation with a scalar drift potential on the periodic domain \(\Omega=[-1,1]^2\):
\[
    \partial_t u
    =
    \nabla \cdot \left( \nabla (u^2) - u \nabla V \right),
    \qquad
    V(x,y)=1-\sin(\pi x)\sin(\pi y).
\]
The initial condition is
\[
    u_0(x,y)
    =
    0.9
    +0.25\cos(\pi x)\cos(\pi y)
    +0.12\sin(2\pi x)\sin(\pi y),
\]
with a small lower cutoff to preserve positivity. The reference solution is computed using a periodic spectral solver on a \(64\times 64\) grid. We use \(\Delta t=10^{-4}\) and evolve the system for \(1000\) time steps.

The neural approximation uses periodic input features
\[
    [\sin(\pi x),\cos(\pi x),\sin(\pi y),\cos(\pi y)],
\]
followed by a fully connected network with two hidden layers of width \(24\), \(\tanh\) activation, and a softplus output map. The network has \(745\) trainable parameters. We compare full EDNN, randomized sparse EDNN using \(25\%\) of the parameters, and a rank-one global SVD EDNN method, where all trainable parameters are put into one single matrix first. The corresponding least-squares update dimensions are \(745\), \(186\), and \(55\), respectively.

\begin{figure}[!htbp]
    \centering
    \includegraphics[width=\textwidth]{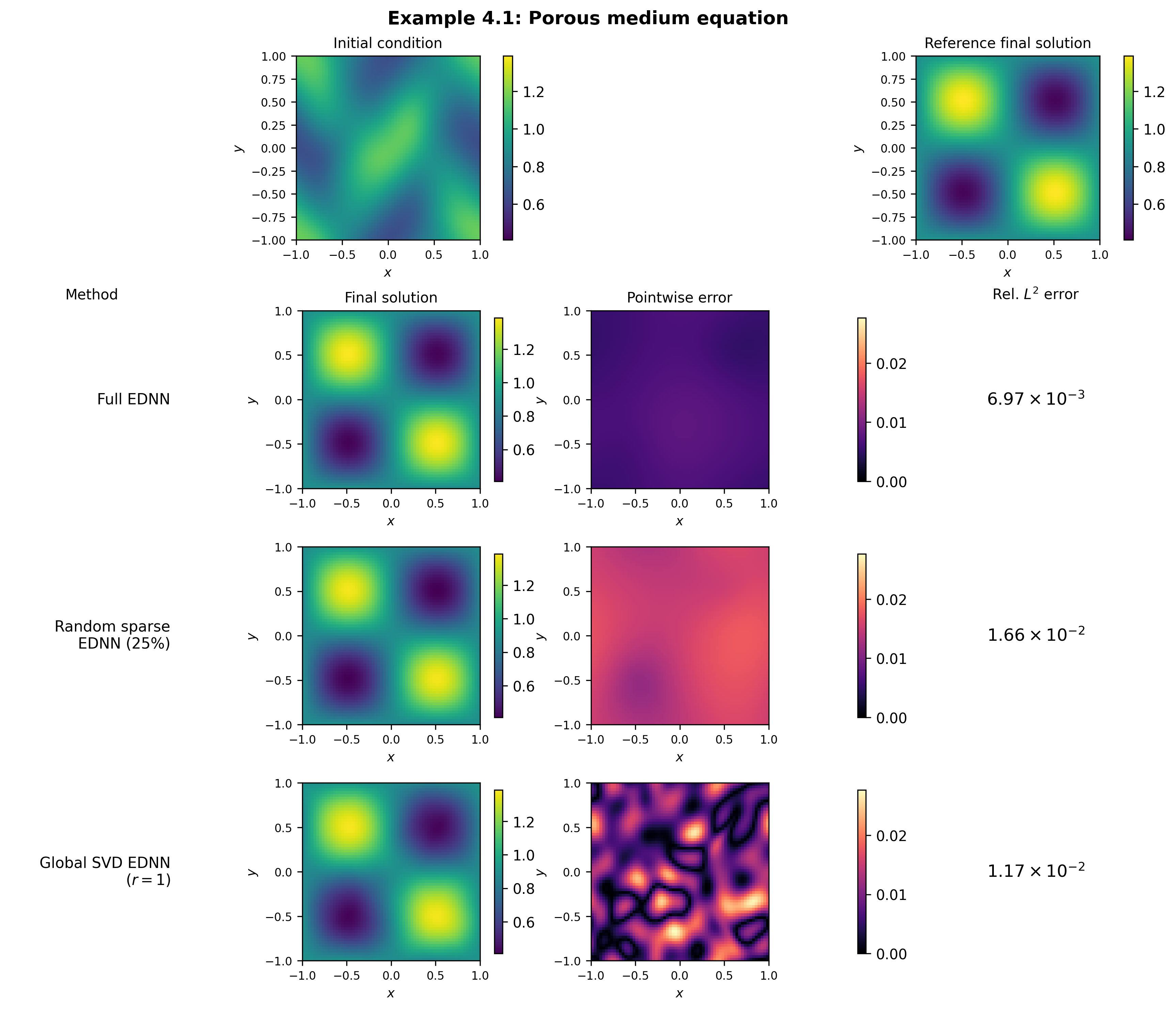}
    \caption{Porous medium equation. Final solution, pointwise error, and final relative \(L^2\) error.}
    \label{fig:porous-medium-summary}
\end{figure}

Figure~\ref{fig:porous-medium-summary} shows that full EDNN gives the smallest final relative \(L^2\) error, \(6.97\times 10^{-3}\). The rank-one global SVD method remains close to the reference solution, with error \(1.17\times 10^{-2}\), while randomized sparse EDNN gives error \(1.66\times 10^{-2}\).

\begin{table}[!htbp]
    \centering
    \caption{Runtime decomposition for the porous medium equation. Times are in seconds. The ``Other'' column includes PDE right-hand-side evaluation, parameter updates, evaluation, diagnostics, and remaining overhead.}
    \label{tab:porous-medium-timing}
    \begin{tabular}{lrrrrr}
        \toprule
        Method & Jacobian & SVD/basis & Linear solve & Other & Total \\
        \midrule
        Full EDNN & 177.22 & 0.00 & 2.96 & 11.41 & 191.59 \\
        Random sparse EDNN (25\%) & 7.95 & 0.00 & 0.92 & 15.21 & 24.08 \\
        Global SVD EDNN (\(r=1\)) & 3.47 & 8.53 & 0.87 & 20.73 & 33.61 \\
        \bottomrule
    \end{tabular}
\end{table}

The timing results show that reducing the update dimension greatly decreases the Jacobian construction cost. Full EDNN forms the full Jacobian \(J_\theta\), while the rank-one global SVD method forms only the reduced matrix \(J_\theta B\). This reduces the Jacobian construction time from \(177.22\) seconds to \(3.47\) seconds.

\FloatBarrier

\subsection{Example 2: One-Dimensional Allen--Cahn Equation}

The second example considers the one-dimensional Allen--Cahn equation on \(\Omega=[-1,1]\) with homogeneous Dirichlet boundary conditions:
\[
    \partial_t u
    =
    \partial_{xx}u
    +
    \frac{u-u^3}{\epsilon^2},
    \qquad
    \epsilon=0.01,
    \qquad
    u(-1,t)=u(1,t)=0.
\]
The initial condition is
\[
    u_0(x)=0.08\sin(\pi x).
\]
We use \(\Delta t=2\times 10^{-6}\) and evolve the system for \(500\) time steps. The reference solution is computed by a finite-difference IMEX Euler solver.

The neural approximation enforces the boundary condition by writing
\[
    u_\theta(x)=(1-x^2)N_\theta(x).
\]
The network input is \([x,\sin(\pi x),\cos(\pi x)]\). The network has three hidden layers of width \(40\), \(\tanh\) activation, and \(3481\) trainable parameters. The EDNN update is computed on \(5000\) interior collocation points. We compare full EDNN, randomized sparse EDNN using \(25\%\) of the parameters, and rank-one layerwise LR-EDNN. The corresponding update dimensions are \(3481\), \(870\), and \(248\).

\begin{figure}[!htbp]
    \centering
    \includegraphics[width=\textwidth]{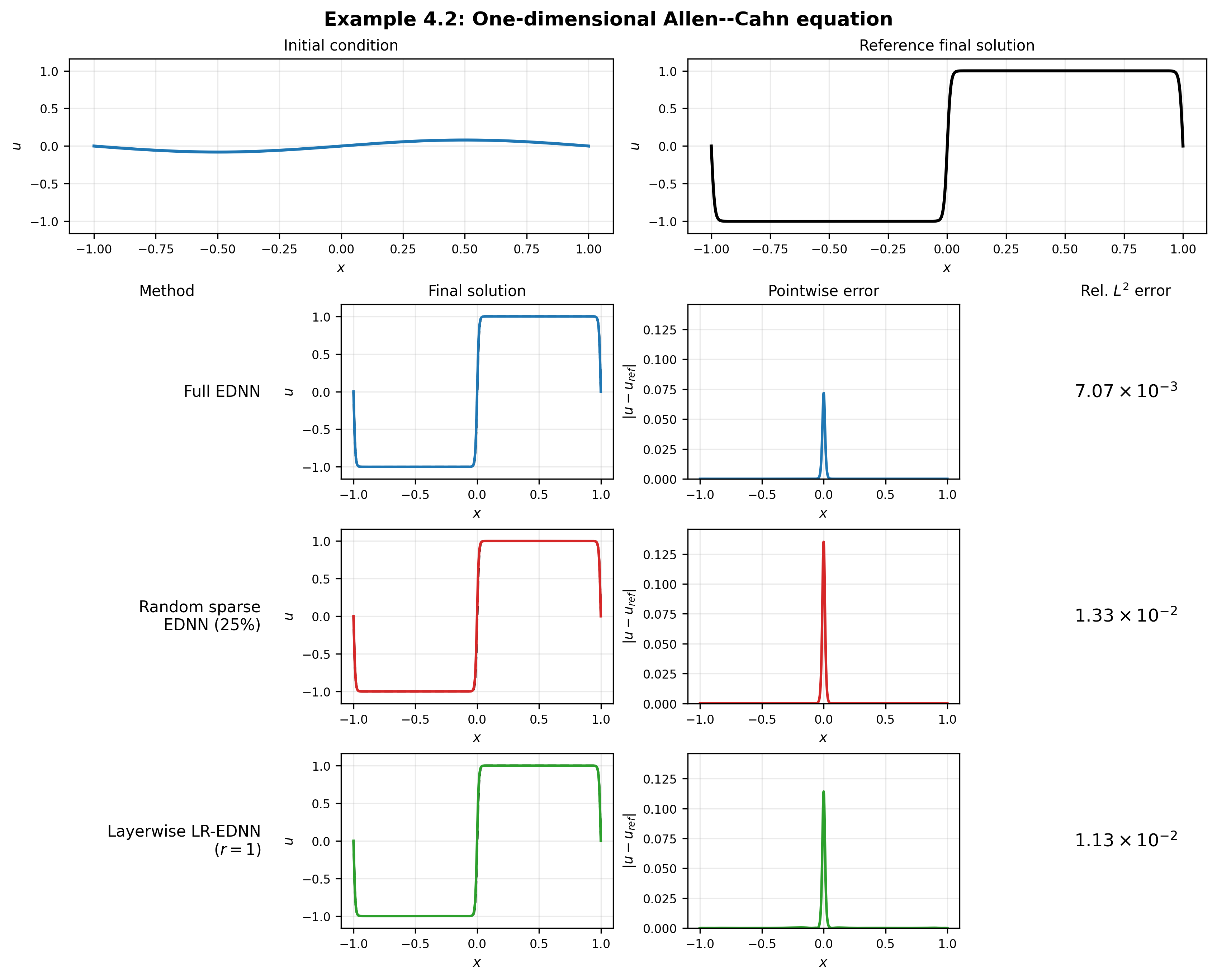}
    \caption{One-dimensional Allen--Cahn equation. Final solution, pointwise error, and final relative \(L^2\) error.}
    \label{fig:allen-cahn-1d-summary}
\end{figure}

Figure~\ref{fig:allen-cahn-1d-summary} shows that full EDNN gives final relative \(L^2\) error \(7.07\times 10^{-3}\). Rank-one layerwise LR-EDNN gives error \(1.13\times 10^{-2}\), while randomized sparse EDNN gives \(1.33\times 10^{-2}\).

\begin{table}[!htbp]
    \centering
    \caption{Runtime decomposition for the one-dimensional Allen--Cahn equation. Times are in seconds.}
    \label{tab:allen-cahn-1d-timing}
    \begin{tabular}{lrrrrr}
        \toprule
        Method & Jacobian & SVD/basis & Linear solve & Other & Total \\
        \midrule
        Full EDNN & 119.53 & 0.00 & 11.96 & 3.51 & 135.00 \\
        Random sparse EDNN (25\%) & 18.86 & 0.00 & 1.67 & 3.64 & 24.16 \\
        Layerwise LR-EDNN (\(r=1\)) & 4.77 & 6.32 & 0.40 & 3.71 & 15.19 \\
        \bottomrule
    \end{tabular}
\end{table}

The update dimension decreases from \(3481\) in full EDNN to \(248\) in rank-one layerwise LR-EDNN. This reduces the Jacobian construction time from \(119.53\) seconds to \(4.77\) seconds and the linear solve time from \(11.96\) seconds to \(0.40\) seconds.

\FloatBarrier

\subsection{Example 3: Two-Dimensional Allen--Cahn Equation}

The third example considers the periodic two-dimensional Allen--Cahn equation on \(\Omega=[-1,1]^2\):
\[
    \partial_t u
    =
    \Delta u
    +
    \frac{u-u^3}{\epsilon^2},
    \qquad
    \epsilon=0.1.
\]
The initial condition is a centered and normalized version of the profile used in Example~1:
\[
    u_0(x,y)
    =
    0.15\,
    \frac{p(x,y)-\overline{p}}{\|p-\overline{p}\|_\infty},
\]
where
\[
    p(x,y)
    =
    0.9
    +0.25\cos(\pi x)\cos(\pi y)
    +0.12\sin(2\pi x)\sin(\pi y).
\]
The reference solution is computed using a periodic Fourier IMEX solver on a \(100\times 100\) grid. We use \(\Delta t=2\times 10^{-4}\) and evolve the solution for \(400\) time steps.

The neural network uses periodic input features, five hidden layers of width \(50\), \(\tanh\) activation, and one scalar output. The network has \(10501\) trainable parameters. We compare full EDNN, randomized sparse EDNN using \(25\%\) of the parameters, and layerwise LR-EDNN with ranks \(r=1\) and \(r=2\). The update dimensions are \(10501\), \(2625\), \(511\), and \(970\), respectively.

\begin{figure}[!htbp]
    \centering
    \includegraphics[width=\textwidth]{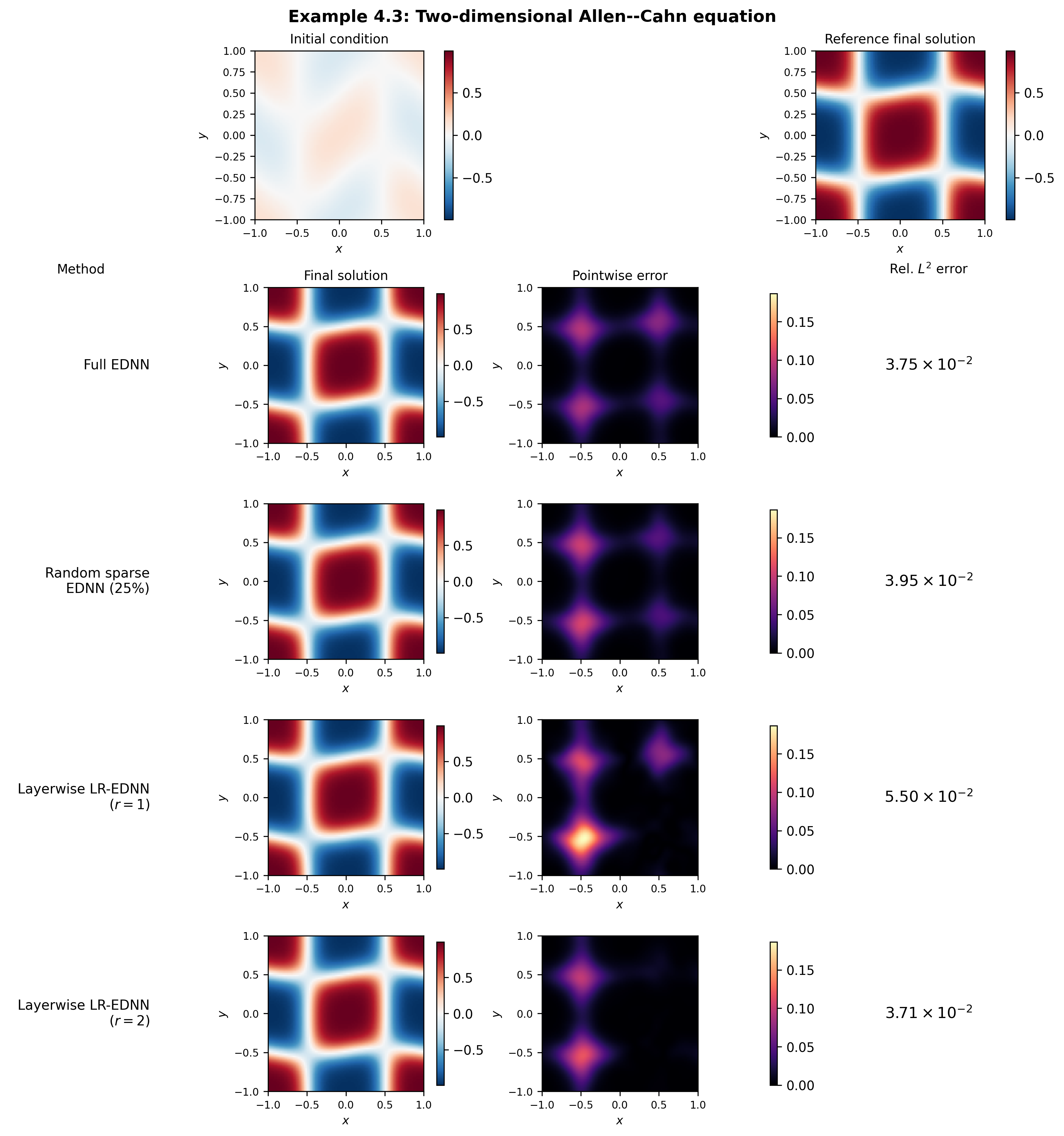}
    \caption{Two-dimensional Allen--Cahn equation. Final solution, pointwise error, and final relative \(L^2\) error.}
    \label{fig:allen-cahn-2d-summary}
\end{figure}

Figure~\ref{fig:allen-cahn-2d-summary} shows that rank-two layerwise LR-EDNN closely matches full EDNN. Full EDNN gives final relative \(L^2\) error \(3.75\times 10^{-2}\), while rank-two LR-EDNN gives \(3.71\times 10^{-2}\). Rank-one LR-EDNN is less accurate, with error \(5.50\times 10^{-2}\), showing that a higher tangent-space rank can be important for more complex two-dimensional dynamics.

\begin{table}[!htbp]
    \centering
    \caption{Runtime decomposition for the two-dimensional Allen--Cahn equation. Times are in seconds.}
    \label{tab:allen-cahn-2d-timing}
    \begin{tabular}{lrrrrr}
        \toprule
        Method & Jacobian & SVD/basis & Linear solve & Other & Total \\
        \midrule
        Full EDNN & 745.22 & 0.00 & 55.40 & 2.94 & 803.56 \\
        Random sparse EDNN (25\%) & 164.44 & 0.00 & 22.02 & 4.97 & 191.43 \\
        Layerwise LR-EDNN (\(r=1\)) & 30.45 & 9.18 & 15.26 & 5.29 & 60.18 \\
        Layerwise LR-EDNN (\(r=2\)) & 58.53 & 14.67 & 15.86 & 5.40 & 94.46 \\
        \bottomrule
    \end{tabular}
\end{table}

Rank-two LR-EDNN solves a \(970\)-dimensional reduced problem instead of the full \(10501\)-dimensional problem. It reduces the total runtime from \(803.56\) seconds to \(94.46\) seconds while maintaining essentially the same final relative \(L^2\) accuracy as full EDNN.

\FloatBarrier

\subsection{Example 4: Two-Dimensional Viscous Burgers Equation}

The final example considers the two-dimensional viscous Burgers system on the periodic domain \(\Omega=[-1,1]^2\):
\[
\begin{aligned}
    \partial_t u + u\partial_x u + v\partial_y u
    &= \nu(\partial_{xx}u+\partial_{yy}u), \\
    \partial_t v + u\partial_x v + v\partial_y v
    &= \nu(\partial_{xx}v+\partial_{yy}v),
\end{aligned}
\]
with viscosity \(\nu=0.025\). The initial condition is
\[
    u_0(x,y)=-\sin(\pi(x+1))\cos(\pi(y+1)),
\]
\[
    v_0(x,y)=\cos(\pi(x+1))\sin(\pi(y+1)).
\]
The reference solution is computed using a periodic Fourier IMEX solver on a \(128\times 128\) grid. We use \(\Delta t=2.5\times 10^{-3}\) and evolve the system for \(200\) time steps.

The neural network uses periodic input features, five hidden layers of width \(70\), \(\tanh\) activation, and two output components. The network has \(20372\) trainable parameters. We compare full EDNN, randomized sparse EDNN using \(25\%\) of the parameters, and layerwise LR-EDNN with ranks \(r=1\) and \(r=6\). The update dimensions are \(20372\), \(5093\), \(712\), and \(3905\), respectively.

\begin{figure}[!htbp]
    \centering
    \includegraphics[width=\textwidth]{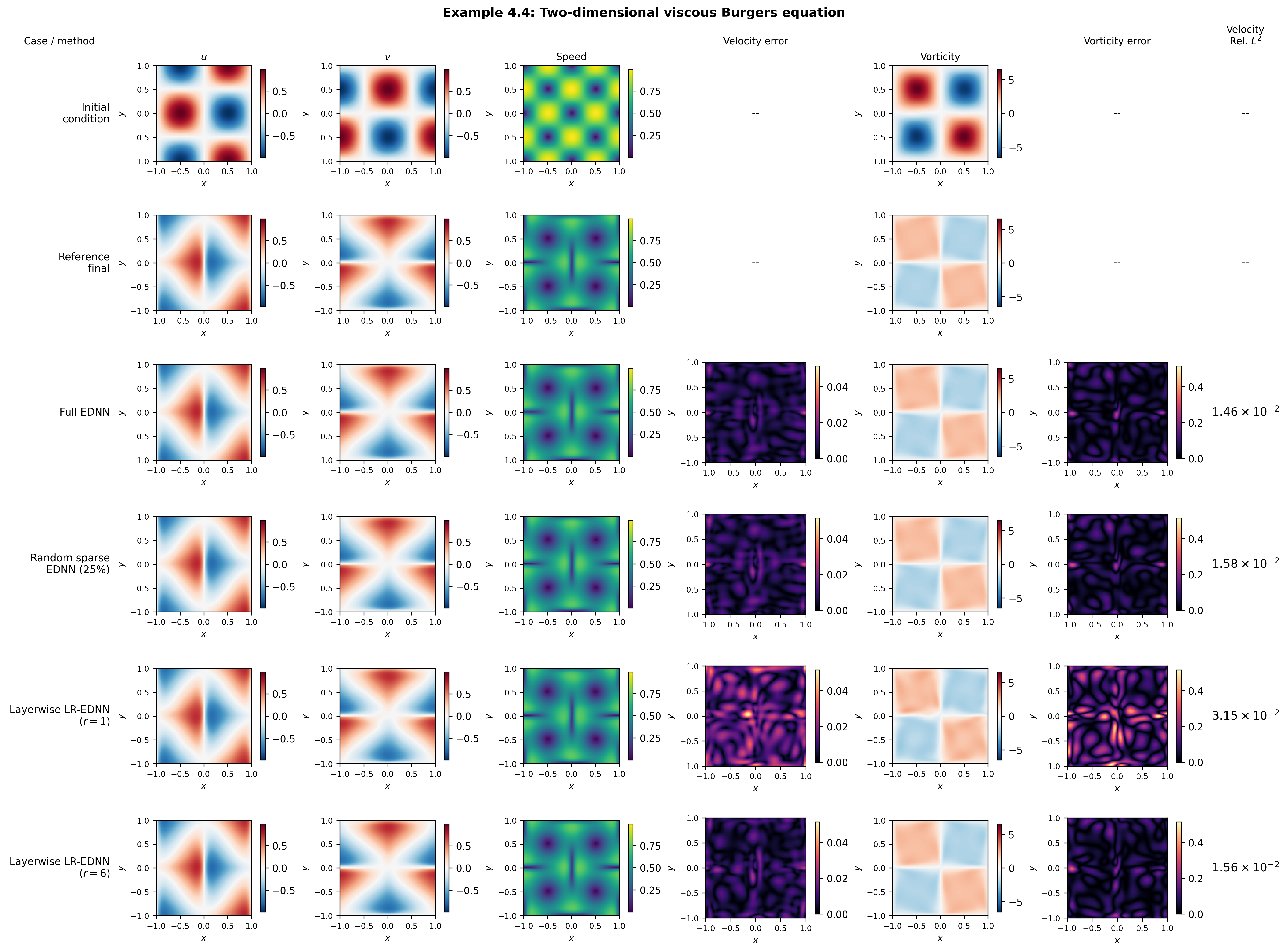}
    \caption{Two-dimensional viscous Burgers equation. Final velocity field, velocity error, vorticity, vorticity error, and final relative \(L^2\) velocity error.}
    \label{fig:burgers-2d-summary}
\end{figure}

Figure~\ref{fig:burgers-2d-summary} reports the final velocity components, speed, velocity error, vorticity, and vorticity error. The final relative \(L^2\) error is computed over both velocity components. Full EDNN gives error \(1.46\times 10^{-2}\), randomized sparse EDNN gives \(1.58\times 10^{-2}\), and rank-six LR-EDNN gives \(1.56\times 10^{-2}\). Rank-one LR-EDNN is less accurate, with error \(3.15\times 10^{-2}\).

\begin{table}[!htbp]
    \centering
    \caption{Runtime decomposition for the two-dimensional viscous Burgers equation. Times are in seconds.}
    \label{tab:burgers-2d-timing}
    \begin{tabular}{lrrrrr}
        \toprule
        Method & Jacobian & SVD/basis & Linear solve & Other & Total \\
        \midrule
        Full EDNN & 3154.64 & 0.00 & 134.25 & 3.09 & 3291.98 \\
        Random sparse EDNN (25\%) & 440.61 & 0.00 & 40.90 & 4.83 & 486.34 \\
        Layerwise LR-EDNN (\(r=1\)) & 60.92 & 6.22 & 10.73 & 4.67 & 82.55 \\
        Layerwise LR-EDNN (\(r=6\)) & 335.46 & 26.04 & 31.04 & 5.09 & 397.62 \\
        \bottomrule
    \end{tabular}
\end{table}

This is the largest experiment in the study. Full EDNN solves over \(20372\) parameter directions and spends more than \(3150\) seconds constructing the Jacobian. Rank-six LR-EDNN reduces the update dimension to \(3905\), lowers the total runtime to \(397.62\) seconds, and gives accuracy close to full EDNN and randomized sparse EDNN. The rank-one result is faster but less accurate, illustrating the accuracy--rank tradeoff in the layerwise low-rank formulation.

\FloatBarrier

%%%%%%%%%%%%%%%%%%%%%%%%%%%%%%%%%%%%%%%%%
%%% Section 5
%%%%%%%%%%%%%%%%%%%%%%%%%%%%%%%%%%%%%%%%%
\section{Result and Discussion}

The numerical results show that LR-EDNN can substantially reduce the computational cost of EDNN while preserving accuracy when the low-rank update space is chosen appropriately. Across the four examples, the main savings come from replacing the full parameter Jacobian \(J_\theta\) with the reduced Jacobian matrix \(J_\theta B\). This reduces both the number of least-squares coordinates and the cost of forming the reduced Jacobian.

\subsection{Accuracy Across Increasing Problem Complexity}

For the smaller porous medium problem in Experiment~4.1, the rank-one global SVD method gives a final relative \(L^2\) error of \(1.17\times 10^{-2}\), compared with \(6.97\times 10^{-3}\) for full EDNN. In Experiment~4.2, rank-one layerwise LR-EDNN gives error \(1.13\times 10^{-2}\), compared with \(7.07\times 10^{-3}\) for full EDNN. These results show that a very low-dimensional update space is already sufficient to capture the dominant dynamics in the simpler scalar problems.

For the larger two-dimensional Allen--Cahn problem in Experiment~4.3, rank matters more. Rank-one LR-EDNN gives error \(5.50\times 10^{-2}\), while rank-two LR-EDNN improves the error to \(3.71\times 10^{-2}\), essentially matching full EDNN at \(3.75\times 10^{-2}\). A similar trend appears in the two-dimensional Burgers system. Rank-one LR-EDNN is the fastest method but gives a larger velocity error, \(3.15\times 10^{-2}\). Increasing the rank to \(r=6\) reduces the error to \(1.56\times 10^{-2}\), close to full EDNN and randomized sparse EDNN.

\subsection{Computational Savings}

\begin{figure}[t]
    \centering
    \includegraphics[width=\textwidth]{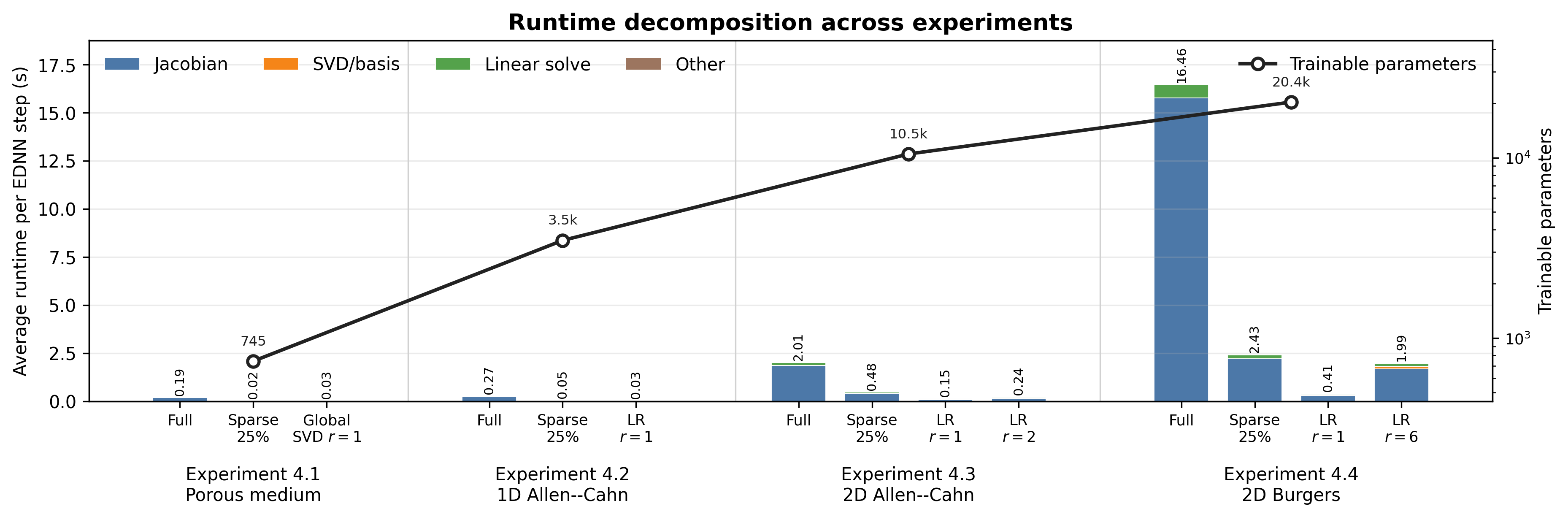}
    \caption{Runtime decomposition across experiments. Each stacked bar reports the average runtime per EDNN step, separated into Jacobian construction, SVD or basis construction, linear solve, and other operations. The black curve shows the number of trainable parameters for each experiment.}
    \label{fig:runtime-decomposition-summary}
\end{figure}

Figure~\ref{fig:runtime-decomposition-summary} summarizes the runtime decomposition across all four experiments. The bars report the average runtime per EDNN step, and the line on the right axis reports the number of trainable parameters. As the model size increases from \(745\) parameters in Experiment~4.1 to \(20372\) parameters in Experiment~4.4, the cost of full EDNN grows rapidly, mainly because forming the full Jacobian \(J_\theta\) becomes expensive.

The benefit of LR-EDNN is most visible in the Jacobian construction time. In Experiment~4.4, full EDNN spends \(3154.64\) seconds forming the full Jacobian, while rank-six LR-EDNN spends \(335.46\) seconds forming \(J_\theta B\). The update dimension is reduced from \(20372\) to \(3905\). Even after including the SVD basis construction cost, the total runtime decreases from \(3291.98\) seconds to \(397.62\) seconds.

\subsection{Layerwise Low-Rank Structure Versus Random Sparsity}

Randomized sparse EDNN also reduces the update dimension, but it does so by selecting individual parameter coordinates. LR-EDNN instead constructs a structured tangent space from the low-rank factors of each layer. This distinction is important. Random sparsity can reduce cost, but it does not use the matrix structure of the neural network layers. The layerwise tangent-space construction retains coherent perturbation directions of the weight matrices and therefore provides a more structured reduced update space.

This difference is visible in the experiments. In Example~2, layerwise LR-EDNN uses \(248\) reduced coordinates, compared with \(870\) selected parameter coordinates for randomized sparse EDNN, while also giving a smaller final relative \(L^2\) error. In Example~3, rank-two LR-EDNN uses \(970\) reduced coordinates and matches full EDNN accuracy more closely than the sparse method, which uses \(2625\) selected parameter coordinates.

\subsection{Effect of Rank}

The rank controls the balance between accuracy and efficiency. A smaller rank gives a lower-dimensional least-squares problem and faster Jacobian construction, but it may not contain enough directions to represent the required parameter velocity. A larger rank increases the cost but improves approximation power.

The two larger examples show this tradeoff clearly. In Example~3, increasing the layerwise rank from \(1\) to \(2\) improves the final relative \(L^2\) error from \(5.50\times 10^{-2}\) to \(3.71\times 10^{-2}\). In Example~4, increasing the rank from \(1\) to \(6\) improves the velocity error from \(3.15\times 10^{-2}\) to \(1.56\times 10^{-2}\). These results suggest that rank-one updates can be effective for simpler scalar dynamics, while more complex or vector-valued systems may require a larger tangent space.

\subsection{Limitations and Practical Considerations}

LR-EDNN introduces an additional SVD or basis construction cost. This cost is visible in every low-rank timing table. However, for the larger examples, the reduction in Jacobian construction and linear solve time is much larger than the SVD overhead. The method is therefore most useful when the full EDNN Jacobian or least-squares system becomes expensive.

The experiments also show that the rank should not be chosen too aggressively. Very low rank can give excellent speedups, but may lose accuracy for more complex dynamics. In practice, the rank should be treated as a controllable accuracy--cost parameter. A small rank can be used first, and then increased when the final error or physical diagnostics indicate that the reduced tangent space is not expressive enough.

Overall, the results support the main claim of this paper: LR-EDNN reduces the computational cost of EDNN by solving in a structured low-rank parameter tangent space, while retaining the ability to approximate the PDE dynamics accurately when the rank is chosen appropriately.

%%%%%%%%%%%%%%%%%%%%%%%%%%%%%%%%%%%%%%%%%
%%% Section 6
%%%%%%%%%%%%%%%%%%%%%%%%%%%%%%%%%%%%%%%%%
\section{Conclusion}

This paper introduced LR-EDNN, a low-rank extension of EDNN designed to reduce the computational cost of neural PDE evolution. The main idea is to restrict the parameter velocity to a structured low-rank tangent space. Instead of forming the full parameter Jacobian \(J_\theta\), LR-EDNN directly constructs the reduced Jacobian matrix \(J_\theta B\), where \(B\) is the global tangent-space parameterization matrix constructed from the current neural network parameters. This avoids materializing the full parameter Jacobian and leads to a much smaller least-squares problem.

A key feature of the method is the layerwise construction of the low-rank tangent space. By applying SVD to each augmented weight matrix, including the bias term, LR-EDNN preserves the matrix structure of the neural network layers. The reduced Jacobian \(J_\theta B\) is then computed by tangent propagation through the network, rather than by materializing \(J_\theta\) and multiplying by \(B\). This makes the method compatible with multi-layer neural networks and avoids the cost of dense full-Jacobian formation.

The numerical experiments show that LR-EDNN can significantly reduce runtime while maintaining good accuracy. Across the porous medium, Allen--Cahn, and viscous Burgers examples, the low-rank methods reduce the update dimension and the Jacobian construction time by a large margin. The larger two-dimensional examples also show that the rank controls the tradeoff between accuracy and efficiency. A small rank gives the fastest computation, while a larger rank can recover accuracy close to full EDNN for more complex dynamics.

Several directions remain for future work. One important question is how to choose the rank adaptively during time evolution. Another direction is to combine LR-EDNN with more advanced iterative solvers, preconditioners, or adaptive collocation strategies. It would also be useful to study larger PDE systems and longer-time integration, where the cost of full EDNN becomes even more restrictive.

Overall, the experiments support the central conclusion that low-rank tangent-space updates provide a practical way to scale EDNN methods. By replacing the full parameter update with a structured reduced update, LR-EDNN keeps the expressive neural representation while making the evolution step substantially more efficient.

\section*{Acknowledgment}
We gratefully acknowledge the support of National Science Foundation (DMS-2533878, DMS-2053746, DMS-2134209, ECCS-2328241, CBET-2347401 and OAC-2311848), and U.S.~Department of Energy (DOE) Office of Science Advanced Scientific Computing Research program DE-SC0023161, the SciDAC LEADS Institute, and DOE–Fusion Energy Science, under grant number: DE-SC0024583.

\bibliographystyle{unsrt}
\bibliography{mybib}

\end{document}